\documentclass{article} 
\usepackage{colm2024_conference}

\usepackage{microtype}
\usepackage{hyperref}
\usepackage{url}
\usepackage{booktabs}
\usepackage{amsfonts}
\usepackage{amsmath}
\usepackage{bbm}
\usepackage{multirow}
\usepackage{caption}
\usepackage{subcaption}
\usepackage{graphicx}
\graphicspath{ {./pics/} }
\usepackage{todonotes}

\title{
Impact of Preference Noise
on the Alignment Performance of Generative Language Models
}

\author{Yang Gao, Dana Alon \& Donald Metzler 
\\
Google Research \\
\texttt{\{gaostayyang, danama, metzler\}@google.com} \\
}

\colmfinalcopy 
\begin{document}

\maketitle

\begin{abstract}
A key requirement in developing Generative
Language Models (GLMs) is to have 
their values aligned with human 
values. 
\emph{Preference-based alignment} is a widely
used paradigm for this purpose, in which
\emph{preferences} over generation pairs are first
elicited from human annotators or AI systems,
and then fed into some alignment techniques,
e.g., 
Direct Preference Optimization.
However, 
a substantial percent
(20 - 40\%)
of the preference pairs
used in GLM alignment are \emph{noisy}, 
and it remains unclear how the noise affects
the alignment performance and how to mitigate
its negative impact.
In this paper, we propose a framework 
to inject desirable amounts and types of
noise to the preferences, and 
systematically study the impact 
of preference noise on the alignment performance 
in two tasks (summarization and dialogue generation).
We find that the alignment performance can be highly
sensitive to the noise rates in the preference data:
e.g., a 10 percentage points (pp) 
increase of the noise rate can lead to
30 pp drop in the alignment performance
(in win rate).
To mitigate the impact of noise,
\emph{confidence-based data filtering} shows significant 
benefit when certain types of noise are present.
We hope our work can help the community better understand and
mitigate the impact of preference noise in GLM alignment.
\end{abstract}

\section{Introduction}
\label{section:introduction}

As the capabilities of \emph{Generative Language Models} (GLMs) keep improving through 
pre-training at a large scale, 
methods for \emph{aligning GLMs with human preferences}, 
i.e., steering GLMs to follow user instructions effectively
and safely, have attracted increasing attention 
\citep{ji2023aialignment}.
%
A widely used paradigm to align GLMs with human values is to 
first collect \emph{binary preferences on generation pairs},
and then use techniques like \emph{Proximal Policy Optimization} 
(a Reinforcement Learning algorithm, \citep{schulman2017ppo}),
\emph{Direct Preference Optimization} (DPO, \citep{rafailov2023dpo}),
or \emph{Sequence Likelihood Calibration} (SLiC, \citep{zhao2023slichf})
to align the GLMs with the collected preferences. 
The binary preferences can be provided by human annotators,
trained \emph{Reward Models} (RMs, \citep{ouyang2022}), 
or \emph{Constitutional AI agents} \citep{bai2022constitutional,lee2023rlaif}.
Preference-based GLM alignment has proven to be 
highly effective in improving the 
safety and usability of GLMs, and hence has been 
used to develop both open-source
\citep{touvron2023llama2,gemma24}
and proprietary \citep{team2023gemini,achiam2023gpt4} GLMs.

However, the binary preferences used in GLM alignment 
are often \emph{noisy},
i.e., containing preferences that disagree with 
the ground truth (e.g., preferences provided by
domain experts).
%
\citet{zheng2023llmjudge} report that 19-37\%
preferences provided by crowd workers 
are noisy. 
%
Similar noise rates are also observed in
preferences provided by RMs and Constitutional AI.
%
Table \ref{table:noise_rates} summarizes the noise 
rates of  preferences used in  recent 
GLM alignment works.
%
%
%
It is generally believed that the lower the noise rates in the 
preferences, the better the final alignment performance
\citep{lee2023rlaif}, 
but it remains unclear what is the \emph{quantitative relation between
noise rates and alignment performance}, and,
furthermore, 
how to \emph{mitigate the negative impact of
 preference noise} on alignment performance.
In this paper, we answer these questions with
systematic empirical studies.

\begin{table}[t]
    \centering
    \small
    \begin{tabular}{l l l l}
    \toprule
    \textbf{Oracle} & \textbf{Task} & \textbf{Noise\%} & \textbf{Reference} \\
    \midrule
    Human 
    & MTBench & 19-37 & \citep{zheng2023llmjudge} \\
    \midrule
    \multirow{5}{*}{Constitutional AI} 
    & MTBench & 15-34 & \citep{zheng2023llmjudge} \\
    & TL;DR & 22 & \citep{lee2023rlaif} \\
    & CBArena & 22-36 & \citep{zheng2023llmjudge} \\
    & AntHH & 27.9-30.9  & \citep{lee2023rlaif} \\
    & MetaHS & 41.4-41.9 & \citep{touvron2023llama2} \\
    \midrule
    \multirow{4}{*}{Reward Models}
    & TL;DR & 21.3-27 & \citep{zhao2023slichf,munos2023nash} \\
    & SHP & 26.3 & \citep{cui2023ultrafeedback} \\
    & MetaHS & 35.5-36.8  & \citep{touvron2023llama2} \\
    & WebGPT & 34.8 & \citep{cui2023ultrafeedback} \\
    \bottomrule
    \end{tabular}
    \caption{
    Preference noise are observed in a wide range of tasks,
    including video game (Atari), QA (MTBench, 
    \textbf{S}tanford\textbf{H}uman\textbf{P}reference),
    Summarization (TL;DR), and Dialogue (WebGPT,
    \textbf{C}hat\textbf{B}ot\textbf{Arena},
    \textbf{Ant}ropic\textbf{H}elpful\textbf{H}armless,
    \textbf{MetaH}elpful\textbf{S}afety).}
    \label{table:noise_rates}
\end{table}

\begin{figure}[t]
    \centering
    \includegraphics[width=0.9\textwidth]{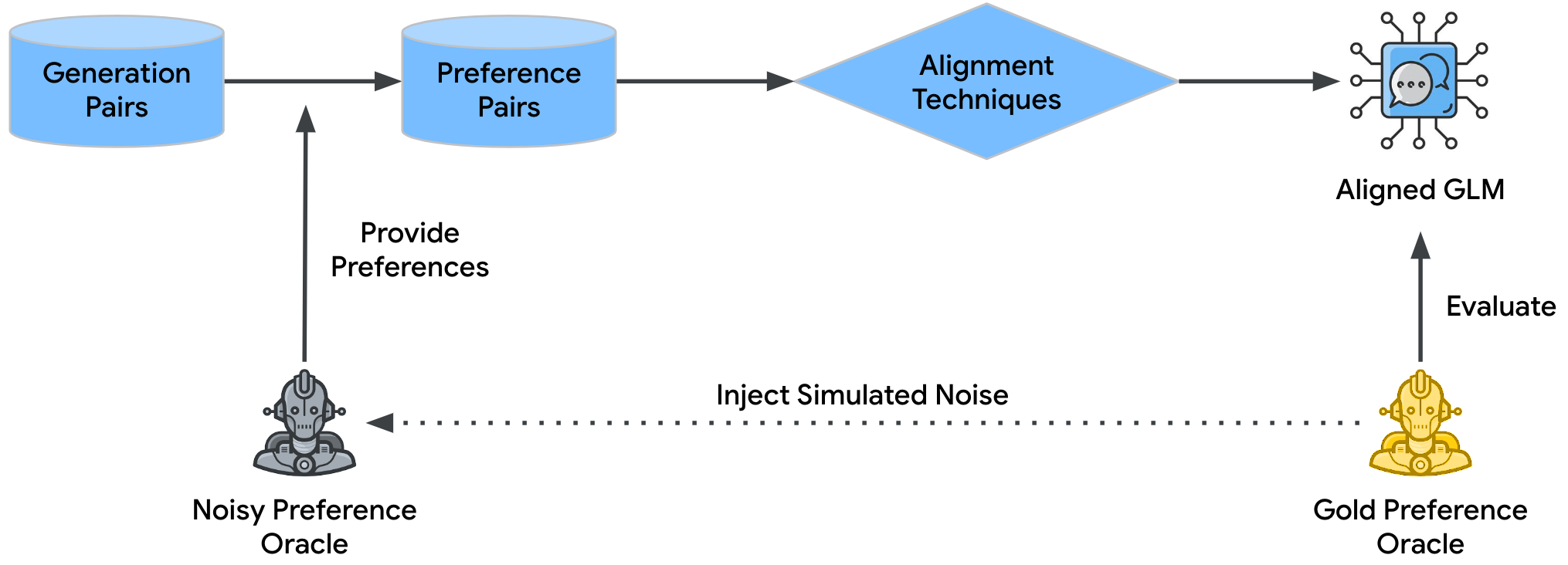}
    \caption{Our framework for evaluating the impact of
    preference noise on GLM alignment.}
    \label{fig:setup}
\end{figure}

There exist frameworks for studying
the influence of preference noise on  alignment
performance in Robotics, e.g., 
the \emph{B-Pref} framework by \citet{lee2021bpref}.
They assume a gold reward model is available, and
design different strategies to \emph{corrupt} the
gold reward model to provide (simulated) noisy preferences
(see \S\ref{section:related_works} for more details).
Although they have been successfully used to benchmark 
and compare different preference-based RL algorithms
in Robotics tasks, they are not applicable for 
GLM alignment, mainly for two reasons:
\textbf{(i)}
some of their noise simulation strategies
are unsuitable for NLP tasks; and
\textbf{(ii)}
the alignment techniques used in Robotics 
(e.g., PREBBLE \citep{lee2021pebble}) 
are different from those used in GLM
(e.g., DPO and SLiC).
To alleviate these problems, we propose a new
framework whose noise-simulation strategies
and alignment techniques are tailored for GLM alignment.
Figure \ref{fig:setup} illustrates the framework.

With the proposed framework, we perform controlled
experiments to study the impact of 
preference noise on  alignment performance
on two tasks, summarization 
\citep{stiennon2020tldr} 
and dialogue generation
\citep{bai2022anthropichh}. 
%
We find that
even with high ($45\%$) noise rates,  
GLM alignment is still beneficial 
(i.e., yielding 50\%+ win rate).
However, alignment performance is also highly sensitive
to noise rates:
a 10 percentage points (pp) increase of noise rates
can lead to 30 pp drop 
in the alignment performance (in terms of win rate).
We also explore different strategies to mitigate
the negative impact of preference noise on 
 alignment performance.
We find that some widely used \emph{regularization} 
methods fail to mitigate the negative impact, 
but \emph{confidence-based data selection} 
can effectively improve  
performance in realistic settings.
%
%
We hope our findings can help  GLM developers
better understand the impact of preference noise
on  alignment performance, and 
that our framework can facilitate the exploration
of more effective and noise-robust alignment methods.

\section{Related Works}
\label{section:related_works}

\paragraph{Preference-based GLM alignment.}
Pairwise preferences are widely used in AI alignment,
because controlled experiments have suggested that 
asking for preferences places a lower cognitive 
burden on the human subjects than asking for absolute 
ratings or categorized labels 
\citep{kendall1948rank,thurstone2017law,kingsley2010preference}.
%
%
In GLM alignment, a common practice is to 
first train an RM from the
human provided preferences, and then use the RM 
to provide reward signals in Reinforcement Learning (RL)
\citep{bohm18better,gao20irj,stiennon2020tldr}. 
Recent methods like DPO \citep{rafailov2023dpo}, 
SLiC \citep{zhao2023slichf}, and 
Identify Preference Optimization (IPO, \citet{azar2023ipo})
go one step further, by eliminating 
the RM training step and directly using 
preference pairs to train the final GLM.
\citet{azar2023ipo}
has shown that 
these methods essentially optimize the
same learning objective, and they differentiate in their
regularization terms and function approximation methods.
Compared to the RM-RL two-stage paradigm, DPO and SLiC
yield stronger performance in multiple NLP applications
with lower computational costs
\citep{chen2024selfplay,yuan2024selfreward}.
%

\paragraph{Learning from noisy data.}
Data used in real-world machine learning applications are
often noisy (c.f., \citet{song2022noisy}), and
%
deep neural models are particularly sensitive to
data noise, as they are prone to overfit to noise patterns in the
training data \citep{DBLP:journals/corr/abs-2011-04406}.
Hence, multiple methods have been proposed
to improve the robustness of the neural models,
mostly falling into four categories
\citep{frenay14noise,song2022noisy}:
\emph{robust neural architectures},
\emph{regularization methods},
\emph{robust loss functions}, and 
\emph{data filtering methods}.
%
In this paper, we stick to the well established 
architecture (Transformers by \citet{Vaswani+2017})
and loss functions (DPO), and explore 
different regularization 
and data selection methods to mitigate
the negative impact of noisy preferences on the alignment 
performance.

The impact of noisy data on RL 
has also been studied.
\citet{lee2021bpref} propose the \emph{B-Pref} benchmark, 
a platform to test the performance and robustness
of preference-based RL algorithms in the face of
different types of preference noise
on various locomotion and robotic manipulation tasks.
To simulate realistic preference noise, they
assume they have access to a gold-standard reward model
$r^*$,
and design five strategies to derive noisy preferences
therefrom, including the 
\emph{stochastic strategy} 
($p(y_0 > y_1) = \sigma[r^*(y_0) - r^*(y_1)]$, where $\sigma$
is the sigmoid function such that
$\sigma(x) = (1+\exp(-x))^{-1}$),
\emph{myopic strategy} (providing preferences only
based on the last part of the presented candidates),
\emph{skipping strategy} (reject to provide preferences
if both candidates are low-quality),
\emph{equally-preferable strategy} (when the quality of
the presented candidates are similar, mark them as a tie),
and \emph{random mistake strategy} (randomly flip the 
correct preference direction with a fixed chance).
We note that some of these strategies are unrealistic
in GLM alignment, e.g., the myopic strategy (users usually
do not judge the quality of texts based on their last parts),
skipping strategy, and equally-preferable strategy
(ties are usually not allowed in annotating text qualities).
Also, they only consider noise from human annotators
but ignore those from AI-based annotators (e.g., 
RLAIF \citep{lee2023rlaif}).
For these reasons, we propose a new set of strategies
for simulating preference noise in GLM alignment
in \S\ref{section:noisy_prefs}.

\section{Preliminaries}
\label{section:preliminaries}

Let $\mathcal{X}$ be the set of all prompts, and 
$\mathcal{Y}_{\mathcal{X}}$ be the set of all possible continuations
for all prompts in $\mathcal{X}$.
%
We assume there exists a gold reward model 
$r^*: \mathcal{X} \times \mathcal{Y}_{\mathcal{X}} \to \mathbb{R}$,
which measures the quality of continuation 
$y \in \mathcal{Y}_{\mathcal{X}}$ for prompt 
$x \in \mathcal{X}$ on some desired aspects (e.g., helpfulness, informativeness,
or harmlessness). 
%
A GLM can be defined as a policy 
$\pi$, 
such that $\pi(y|x)$ is the probability of
generating $y$ for the input prompt $x$.
The objective of \emph{GLM alignment} is to
find the optimal policy that can 
maximize the expected gold reward value 
while minimizing the divergence from a reference policy:
\begin{align}
\max_\pi  \sum_{x\in\mathcal{X},
y\sim\pi(\cdot|x)} [r^*(y|x)] - 
\beta\mathbb{D}_{KL}[\pi_\theta(y|x) || 
\pi_\text{sft}(y|x)],
\label{eq:glm_alignment}
\end{align}
where $\beta$ is a hyperparameter, 
$\mathbb{D}_{KL}$ is the Kullback–Leibler divergence,
and $\pi_\text{sft}$ is the reference policy. 

In practice, 
we cannot 
optimize Eq. \eqref{eq:glm_alignment} directly,
because the gold reward $r^*$ is usually inaccessible,
and the summation operation is prohibitively expensive.
Multiple algorithms have been proposed to obtain 
(approximate) solutions for the objective 
(see \S\ref{section:related_works}). 
In this work, we use
DPO \citep{rafailov2023dpo} because of its strong performance
and lower computational cost compared
to other methods (e.g., PPO). The loss
function in DPO is:  
\begin{align}
\mathcal{L}(\pi_\theta)  =  
-\mathbb{E}_{(x,y_w,y_l) \sim \mathcal{D}} \{ 
\log\sigma[
\beta \log(\frac{\pi_\theta(y_w|x)}{\pi_\text{sft}(y_w|x)}) 
-
\beta \log(\frac{\pi_\theta(y_l|x)}{\pi_\text{sft}(y_l|x)})
]\},
\label{eq:dpo_slic}
\end{align}
where $\pi_\theta$ is the learnable policy parameterized
by $\theta$, 
$\sigma$ is the sigmoid function, 
and $\mathcal{D} = \{(x^i, y^i_w, y^i_l)\}_{i=1}^{n}$
is the training dataset consisting of $n$ data entries.
Each data entry in $\mathcal{D}$ consists of a prompt 
$x^i \in \mathcal{X}$ and
two continuations 
$y^i_w, y^i_l \in \mathcal{Y}_{\mathcal{X}}$, 
such that $y^i_w$ is preferred over $y^i_l$.

It has been proven that if $\mathcal{D}$ is sufficiently large
and all pairs in $\mathcal{D}$ are noise-free 
(i.e., $r^*(y^i_w) > r^*(y^i_l)$ for $i=1, \cdots, n$),
the policy learned by DPO is (near-)optimal with respect to
Eq. \eqref{eq:glm_alignment} \citep{rafailov2023dpo,azar2023ipo}.
However, in practice, some preferences in $\mathcal{D}$ 
can be \emph{noisy},
i.e., different from the preference direction induced 
by the gold reward model. 
In this work, we remove the (strong) noise-free assumption 
on $\mathcal{D}$, but instead introduce different rates
and types of noise to $\mathcal{D}$ (in \S\ref{section:noisy_prefs}) 
and empirically study 
their impact on the quality of $\pi_\theta$ (in \S\ref{section:result_vanilla}).


To measure the alignment performance (i.e., measure
the performance of $\pi_\theta$),
we compute the \emph{win rate} between 
$\pi_\theta$ and 
$\pi_{\text{sft}}$:
\begin{align}
w = \frac{1}{|\mathcal{X}_{\text{test}}|} 
\sum_{x \in \mathcal{X}_{\text{test}}} 
\mathbbm{1}[r^*(x, y_{\pi_{\theta}}) > 
r^*(x, y_{\pi_\text{sft}})],
\label{eq:win_rate}
\end{align}
where $\mathcal{X}_{\text{test}} \subset \mathcal{X}$
is a held-out test prompt set, and
$y_{\pi_\theta}$ and $y_{\pi_\text{sft}}$ are 
generations sampled from
$\pi_{\theta}$ and $\pi_\text{sft}$, respectively.

\section{Noisy Preferences}
\label{section:noisy_prefs}

The preferences are often \emph{noisy},
i.e., disagree with the preference directions 
induced by the gold reward model $r^*$.
Inspired by past works 
\citep{lee2021bpref} (see \S\ref{section:related_works}
for more discussions),
we consider three \emph{oracles} to provide
different types of noisy preferences.

\begin{itemize}
\item 
\textbf{Random Noise Oracle}. When presented with 
a prompt $x \in \mathcal{X}$ and a pair of 
responses $y_w, y_l \in \mathcal{Y}_\mathcal{X}$,
the oracle has $(100-n)$\% chance to return the 
correct preference 
(i.e., $r^*(y_w|x) > r^*(y_l|x)$), 
but has $n$\% chance to
return the incorrect/flipped preferences. 
We can control the noise rate 
of this oracle by adjusting the value of $n$.

\item 
\textbf{Stochastic Noise Oracle}. 
For a prompt $x$ and two responses $y_w, y_l$,
Stochastic Noise Oracle prefers $y_w$ over $y_l$
with the probability $\sigma[(r^*(y_w)-r^*(y_l))/\gamma]$, 
where $\sigma$ is the sigmoid function, and
$\gamma \in \mathbb{R}^+$ is the temperature hyperparameter.
We can control the noise rate 
by tuning the $\gamma$ value:
The higher the $\gamma$ value, the more unpredictable
the oracle is, and hence more noisy the preferences will be.

\item
\textbf{Gaussian Noise Oracle}.
Stochastic Noise Oracle requires access to the
gold reward model, which is infeasible in practice.
Gaussian Noise Oracle, instead, only requires the access to 
an approximated reward model $r'$, such
that $r'(y|x) = r^*(y|x) + \epsilon$, where $\epsilon$
is the \emph{noise term} drawn from a Gaussian distribution 
$\mathcal{N}(\mu, \delta^2)$. The preference directions
are then derived from the approximated reward $r'$.
%
With $r'$, the probability of Gaussian Noise Oracle
prefers $y_w$ over $y_l$ is:
\begin{align*}
 p(y_w > y_l | x)  = & \mathbbm{1}[r'(y_w|x) - r'(y_l)]  \\ 
= & \mathbbm{1}[(r^*(y_w|x) + \epsilon_w) - (r^*(y_l|x) + \epsilon_l)] \\
= & \mathbbm{1}[(r^*(y_w|x) - r^*(y_l|x)) + 
(\epsilon_w - \epsilon_l)].
\end{align*}
Since both $\epsilon_w$ and $\epsilon_l$ are drawn from 
the same Gaussian distribution $\mathcal{N}(\mu, \delta^2)$, 
$\epsilon_w - \epsilon_l$ is a random variable drawn from
$\mathcal{N}(0, 2\delta^2)$.
Hence, the noise rate of Gaussian Noise
Oracle can be adjusted by tuning the value of $\delta$.
\end{itemize}

We believe the three strategies cover some widely observed
noise types in preferences. 
For example, Stochastic Noise Oracle is 
also known as \emph{Boltzmann rational}
\citep{ziebart2008maximum,jeon2020reward,gao20irj}
and widely used for simulating noise in 
human-provided preferences caused by \emph{aleatoric uncertainty}
\citep{hullermeier2021aleatoric}.
Gaussian Noise Oracle simulates the noise caused 
by the \emph{epistemic uncertainty}, i.e., 
the RMs fail to accurately approximate the 
human's preferences.
Random Noise Oracle simulates the random mistakes
observed in both human-provided
\citep{lindner2022humans} and heuristic-based
preferences \citep{chen2024selfplay}.
We note that in reality, multiple types of 
noise can co-exist in the preference pairs;
however, to ease analyses, in this paper,
we assume there exist at most one type of noise
in the preferences. We leave mixed-type preference
noise for future work.

\section{Experimental Setup}
\label{section:setup}

\paragraph{Tasks.}
We consider preference-based GLM alignment on two tasks:
Reddit TL;DR \citep{stiennon2020tldr}
and Anthropic-Helpful
\citep{bai2022anthropichh}. 
Reddit TL;DR 
has two subsets, a SFT set and a preference set.
In its preference set,
each data entry consists of a document $x$ and two
candidate summaries $y_w, y_l$ for $x$.
It has 93k/53k/33k data entries in train/validation/test splits.
Anthropic-Helpful is a subset of the AnthropicHH dataset,
in which each data entry contains
the dialogue history between a human and an AI assistant ($x$),
and two candidate responses ($y_w, y_l$).
It has 161k/9k data entries in train/test splits, and
we separate out 1k randomly-sampled entries from 
the train set as the validation set.

\paragraph{Generative Language Model.}
For each task,
we fine-tune a T5-Large (770M parameters) model to obtain 
the initial GLM $\pi_\text{sft}$. 
For TL;DR, we use the SFT subset in Reddit TL;DR as the SFT training
data, which has 117k/6k/6k examples in train/validation/test splits.
For Anthropic-Helpful, we use all the preferred responses
in the dataset as the SFT training data.
All the hyperparameters used in SFT are the same as 
in \citep{liu2023statistical}.

\paragraph{Gold Reward Model $r^*$.}
For each task (TL;DR and Anthropic-Helpful), 
we train a T5-XXL (11B parameters) \citep{raffel2020t5x} 
model with the respective preference pairs 
to build the gold reward model $r^*$.
In line with \citep{zhao2023slichf}, 
we format the input to the model with the prompt: \texttt{[CONTEXT] \{$x$\} [RESPONSE] \{$y$\}}, 
and use the logit of the token \texttt{1} as a point-wise 
score for the reply.

%

\paragraph{Noisy Preference Oracles.}
Based on the gold reward model $r^*$ described above, we
build the noisy preference oracles as described in 
\S\ref{section:noisy_prefs}. 
The noise rate of different
noisy preference oracles can be controlled by tuning their respective
hyperparameters: $n$ (noise rate) for 
Random Noise Oracle, $\gamma$ (temperature) for Stochastic Oracle,
and $\delta$ (standard deviation) for Gaussian Noise Oracle. 
To decide the exact values of the hyperparameters for
a target noise rate (e.g., 20\%),  
we randomly sample 1k examples from each train set,
and increase the hyperparameters with a small step ($0.01$)
until the target noise rate is reached.
The final hyperparameter values are presented in 
Table \ref{table:oracle_hyperp}.


\begin{table}[h]
    \centering
    \begin{tabular}{l l l l l l l l l l l l}
    \toprule
     & & 5\% & 10\% & 15\% & 20\% & 25\% & 
     30\% & 35\% & 40\% & 45\% & 50\% \\
     \midrule
     \multirow{2}{*}{TL;DR} & $\gamma$ & 0.2 & 0.4 & 0.65 & 0.9 & 1.25 &
     1.75 & 2.50 & 3.90 & 8.0 & 100 \\
      & $\delta$ & 0.34 & 0.70 & 1.09 & 1.55 & 2.15 &
     2.90 & 4.05 & 6.50 & 13.0 & 100 \\
     \midrule
     \multirow{2}{*}{Anthropic} & $\gamma$ &
     0.11 & 0.22 & 0.36 & 0.53 & 0.75 &
     1.06 & 1.54 & 2.45 & 4.95 & 100 \\
     & $\delta$ & 0.18 & 0.40 & 0.62 & 0.89 & 1.25 &
     1.75 & 2.49 & 4.00 & 8.75 & 100 \\
    \bottomrule
    \end{tabular}
    \caption{
    Hyperparameters for the Stochastic Noise ($\gamma$) 
    and Gaussian Noise Oracle ($\delta$) at each target noise
    rate (column). The hyperparameter for Random Noise Oracle 
    ($n$; see \S\ref{section:noisy_prefs}) 
    is omitted, as it equals the corresponding target noise rate.}
    \label{table:oracle_hyperp}
\end{table}

\paragraph{Generation Pairs.}
In line with \citet{liu2023statistical},
we use the trained GLM
$\pi_\text{sft}$ to sample responses for each prompt, and pair up 
the sampled responses to build the Generation Pairs dataset 
(see Fig. \ref{fig:setup}). 
%
%
For each prompt $x$, we sample eight generations 
from $\pi_\text{sft}(\cdot | x)$
with temperature $0.7$,
and randomly group them into four pairs.
The pairs are then presented to the noisy preference oracles
to build the 
Preference Pairs dataset (see Fig. \ref{fig:setup}).

\paragraph{Other Hyperparameters.}
We choose the hyperparameters by following the choices
made in \citep{rafailov2023dpo,liu2023statistical}:
in training the gold reward model and GLM $\pi_\text{sft}$,
we use batch size 32 and learning rate 1e-5 with
Adafactor optimizer \citep{shazeer2018adafactor};
in the alignment training stage 
(Eq. \eqref{eq:dpo_slic}), we use
$\beta=0.5$, and dropout rate 0.1.
Later in \S\ref{section:result_mitigation}, we will explore
different values for the regularization weights
($\beta$ and dropout rate) to study their effectiveness in
mitigating the negative effect of noisy preferences.

\section{Impact of Noise Rates on Alignment Performance}
\label{section:result_vanilla}

\begin{figure}
\centering
    \begin{subfigure}[b]{0.48\textwidth}
         \centering
         \includegraphics[width=\textwidth]{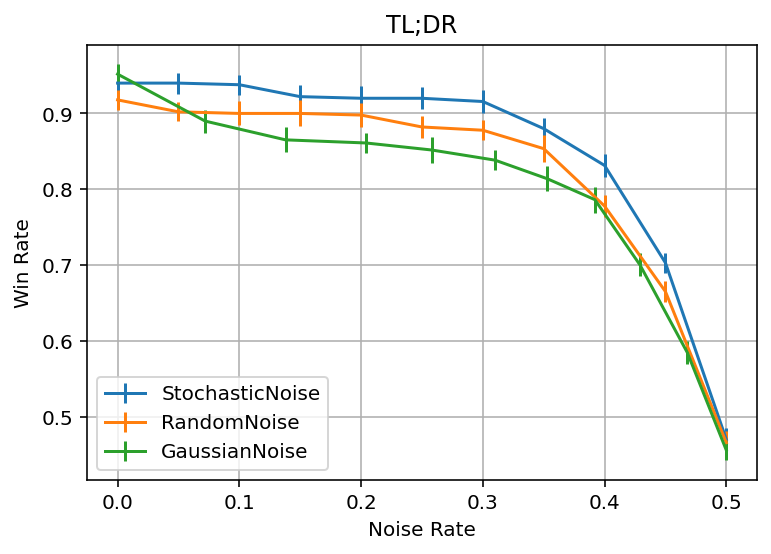}
     \end{subfigure}
    \begin{subfigure}[b]{0.48\textwidth}
         \centering
         \includegraphics[width=\textwidth]{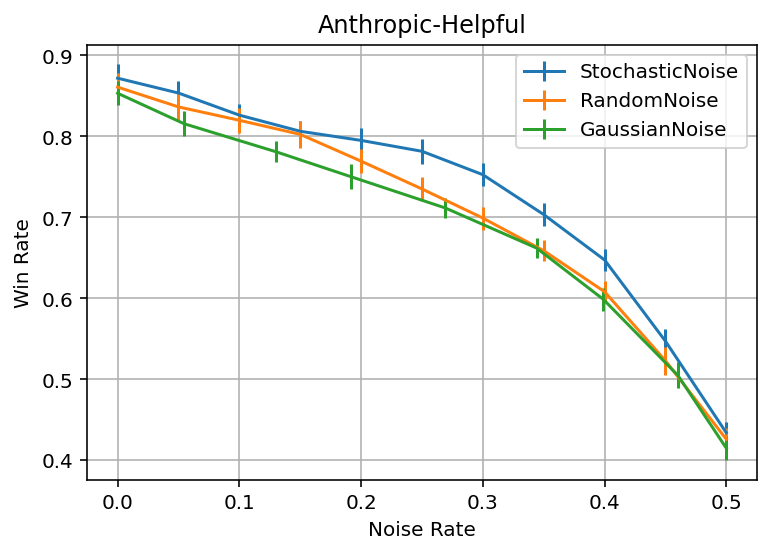}
     \end{subfigure}
     \caption{
Influence of the noise rate (x-axis) on the alignment performance (in terms
of the win rate; y-axis).
Error bars: 95\% confidence intervals, 
computed with double-tailed t-test on 5-10 repeated
experiments with different random seeds.
}
\label{fig:results_vanilla}
\end{figure}

Fig. \ref{fig:results_vanilla} presents how the 
alignment performance 
changes with the growth of the noise rates. 
We make the following observations.

\begin{itemize}
\item
\textbf{Alignment performance drops with more noise
in preferences.}
This applies to all types of noise and 
both tasks we have considered. 
Also, we note that the alignment performance
drops more quickly with the increase of the noise rates:
When the noise rates are below 0.3,
an increase of 10 percentage points (pp)
in noise rate yields less than 10pp 
drop in the alignment performance;
however, when the noise rates are higher than 
0.4, 10pp increase in noise can
yield 20-30 pp loss in performance.

\item
\textbf{Different types of noise cause similar harm.}
At the same noise rate, the alignment performance of the
three different noise types do not have significant differences,
suggesting that it is the noise rate rather than the 
noise type that decides the alignment performance.

\item
\textbf{Alignment is beneficial even 
with highly noisy preferences.}
In both tasks, the win rate is above $0.5$ 
even with noise rate at $0.45$.
This observation reaffirms the effectiveness of 
alignment training \citep{touvron2023llama2}, and explains why even
highly noisy preferences are used in alignment training in practice 
(see Table \ref{table:noise_rates}).
But it is also worth noting that 
when the preferences are completely 
random (i.e., noise rate at 50\%),
the win rate drops below 0.5, suggesting that alignment is detrimental
with random preferences.
\end{itemize}

\section{Mitigate the Negative Impact of Preference Noise}
\label{section:result_mitigation}
In this section, we explore two popular strategies 
to mitigate the negative impact of preference noise: 
\emph{regularization} in \S\ref{subsec:regularization}
and \emph{data filtering} in \S\ref{subsec:data_filter}.

\subsection{Regularization}
\label{subsec:regularization}

We consider two methods to strengthen regularization:
increasing the weight of the KL divergence loss ($\beta$
in Eq. \eqref{eq:dpo_slic}), 
and increasing the dropout rate.

\begin{figure}
\centering
\begin{subfigure}[b]{0.32\textwidth}
 \centering
 \includegraphics[width=\textwidth]{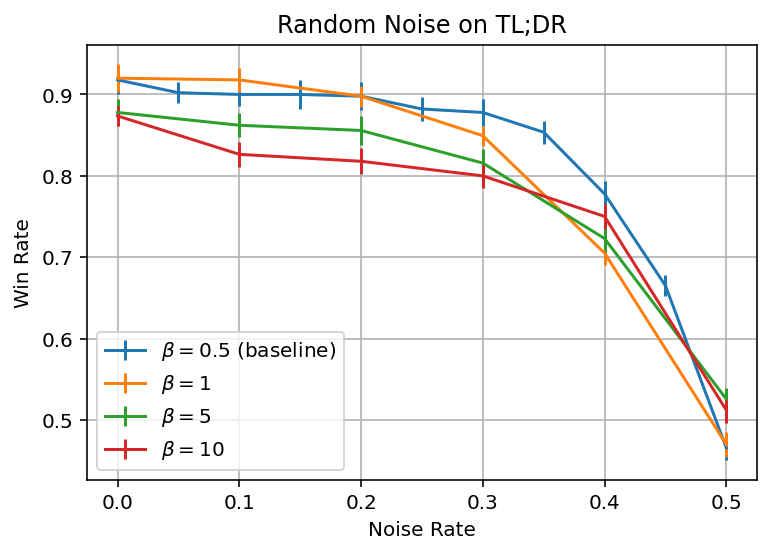}
\end{subfigure}
\begin{subfigure}[b]{0.32\textwidth}
 \centering
 \includegraphics[width=\textwidth]{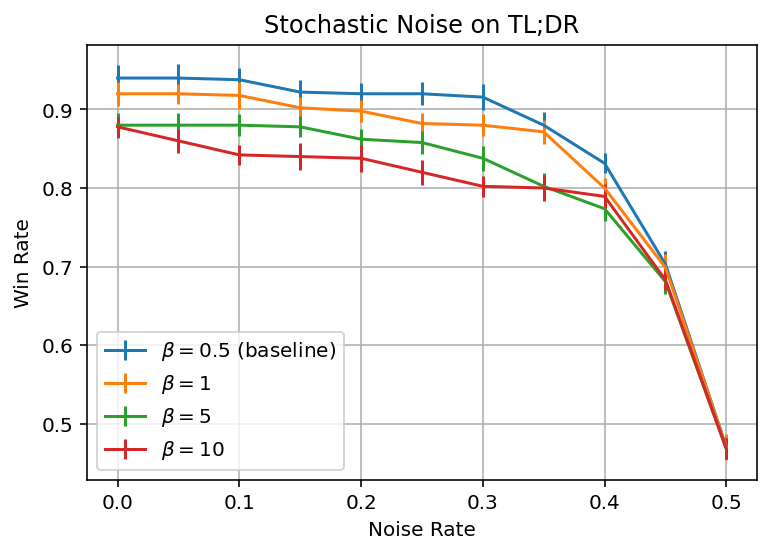}
\end{subfigure}
\begin{subfigure}[b]{0.32\textwidth}
 \centering
 \includegraphics[width=\textwidth]{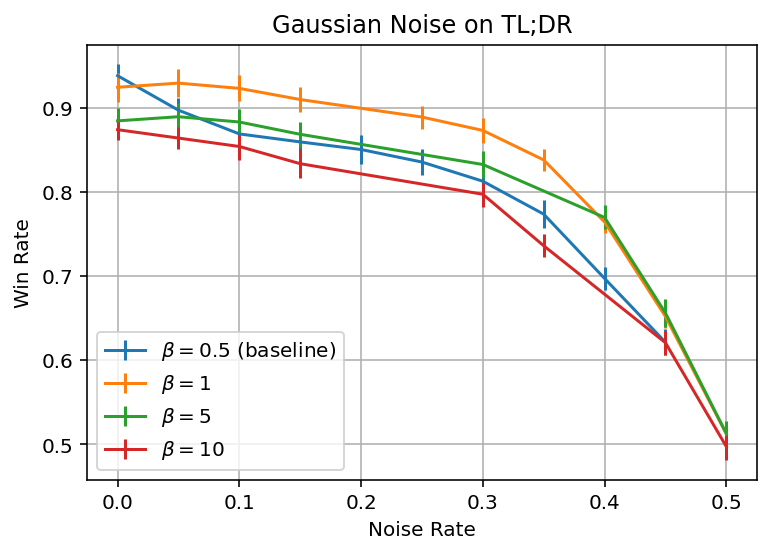}
\end{subfigure}
\begin{subfigure}[b]{0.32\textwidth}
 \centering
 \includegraphics[width=\textwidth]{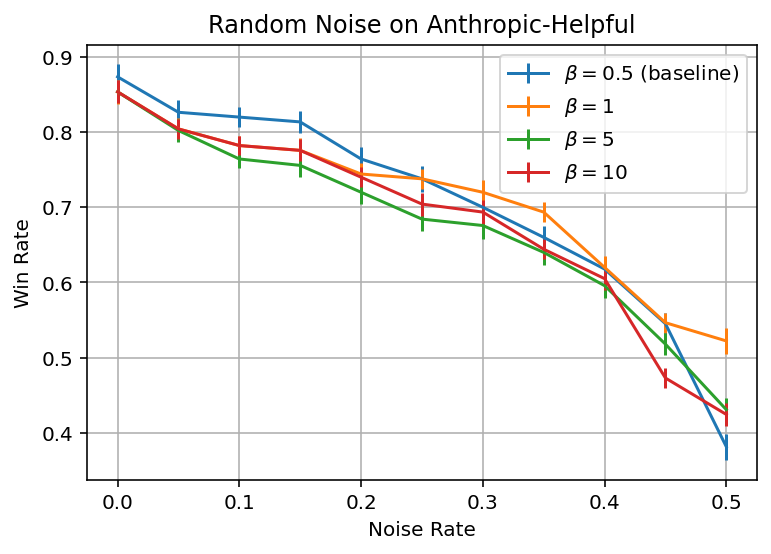}
\end{subfigure}
\begin{subfigure}[b]{0.32\textwidth}
 \centering
 \includegraphics[width=\textwidth]{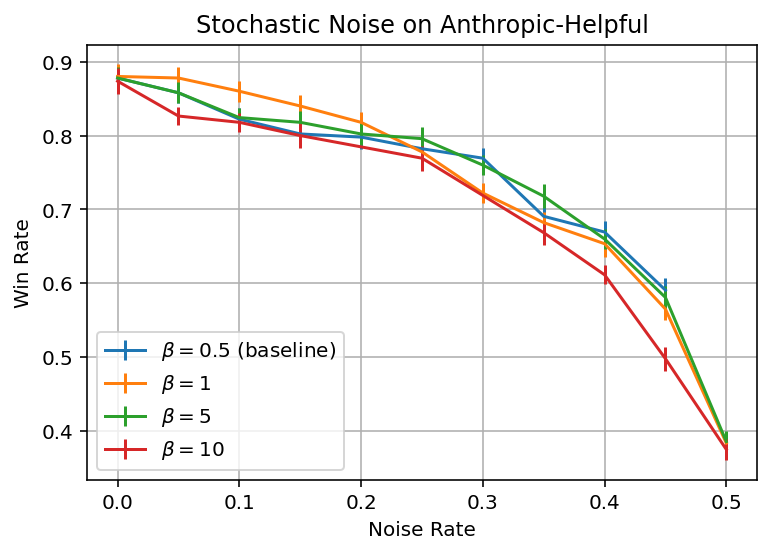}
\end{subfigure}
\begin{subfigure}[b]{0.32\textwidth}
 \centering
 \includegraphics[width=\textwidth]{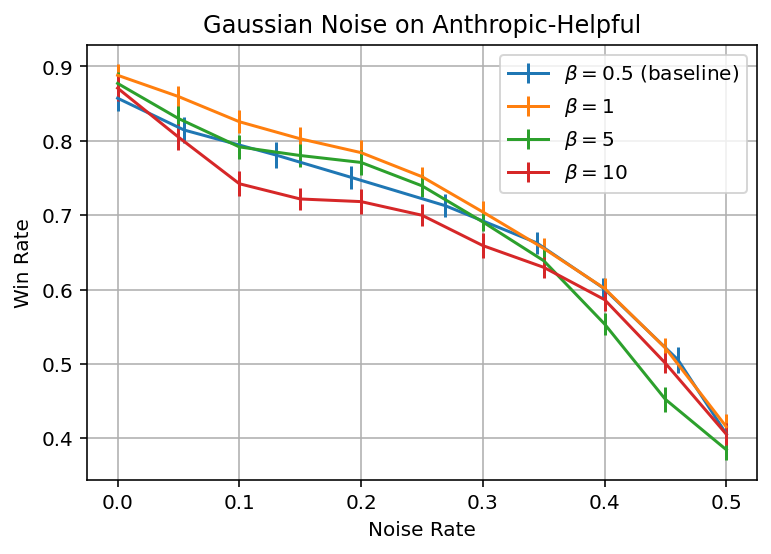}
\end{subfigure}
\caption{Alignment performance with different KL regularization weights $\beta$.}
\label{fig:results_kl}
\end{figure}

Fig. \ref{fig:results_kl} presents the alignment performance with 
different strengths of KL regularization. 
In general, we find that stronger KL regularization
fails to mitigate the negative impact of preference noise. 
In some cases
(e.g., Random and Stochastic Noise in TL;DR), higher strength of KL regularization even hurts the performance. 
%
Our finding reaffirms the
limitations of KL-based regularization
in DPO/SLiC \citep{azar2023ipo}.

\begin{figure}
\centering
\begin{subfigure}[b]{0.32\textwidth}
 \centering
 \includegraphics[width=\textwidth]{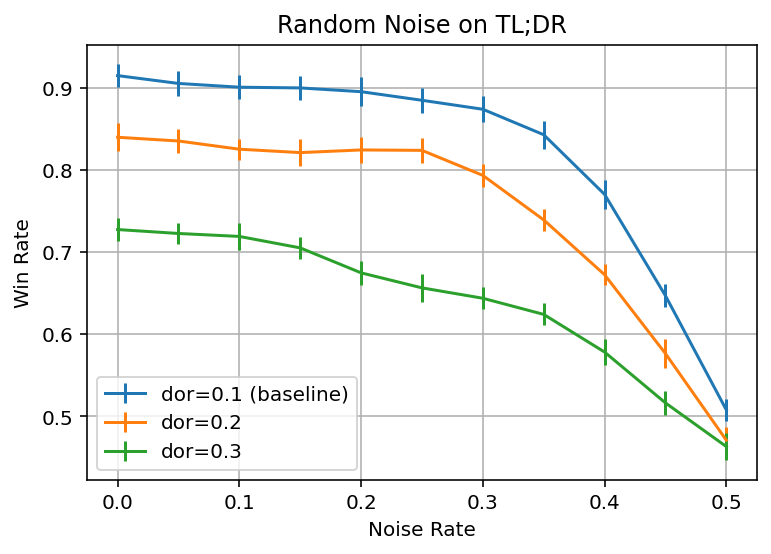}
\end{subfigure}
\begin{subfigure}[b]{0.32\textwidth}
 \centering
 \includegraphics[width=\textwidth]{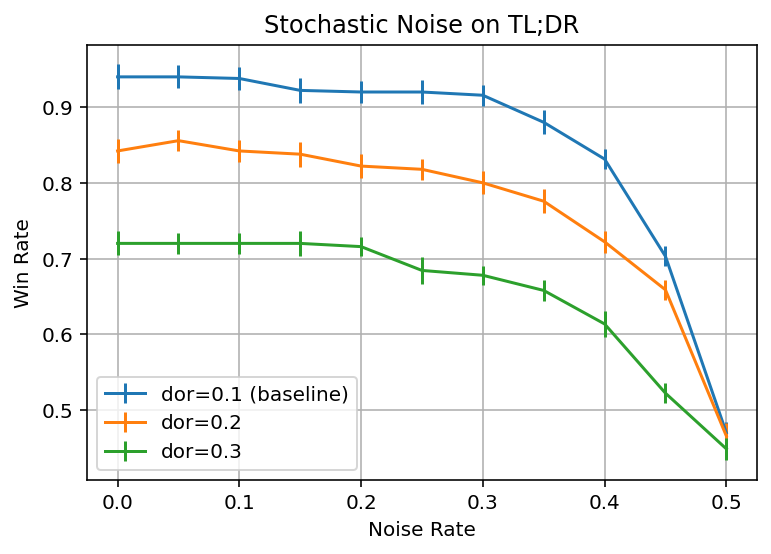}
\end{subfigure}
\begin{subfigure}[b]{0.32\textwidth}
 \centering
 \includegraphics[width=\textwidth]{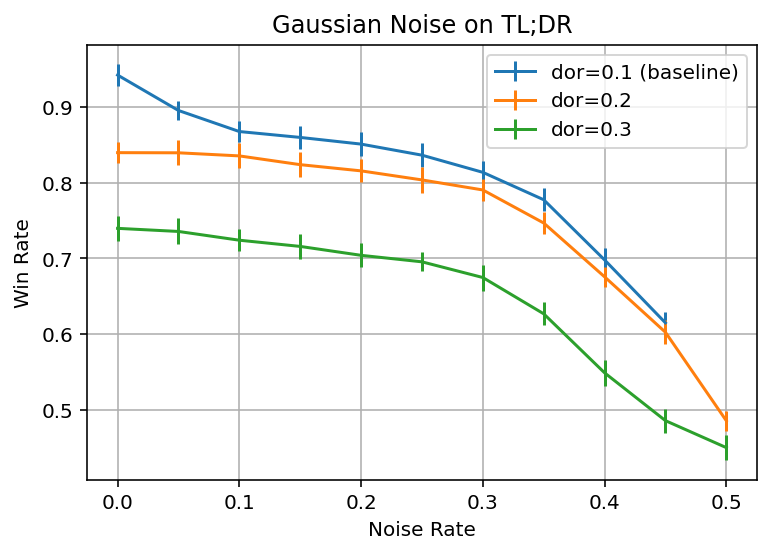}
\end{subfigure}
\begin{subfigure}[b]{0.32\textwidth}
 \centering
 \includegraphics[width=\textwidth]{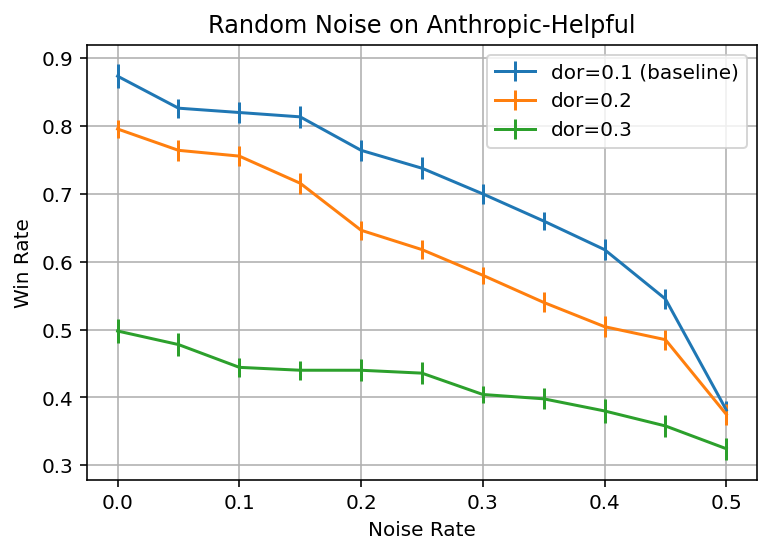}
\end{subfigure}
\begin{subfigure}[b]{0.32\textwidth}
 \centering
 \includegraphics[width=\textwidth]{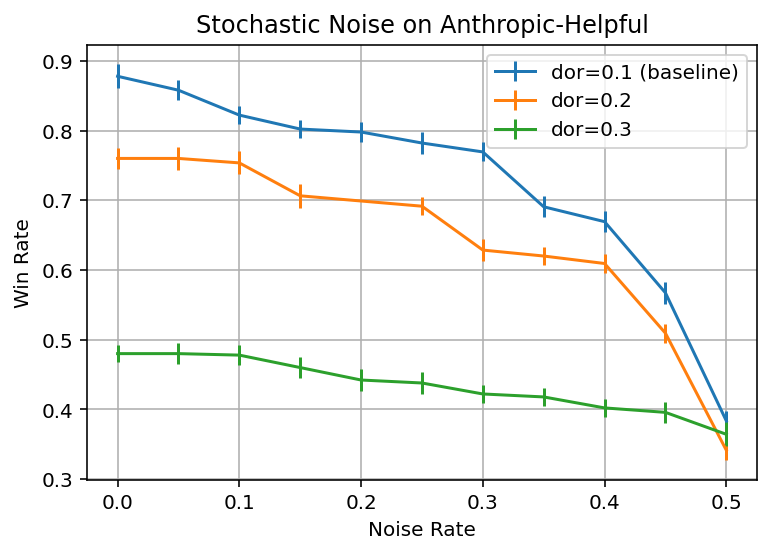}
\end{subfigure}
\begin{subfigure}[b]{0.32\textwidth}
 \centering
 \includegraphics[width=\textwidth]{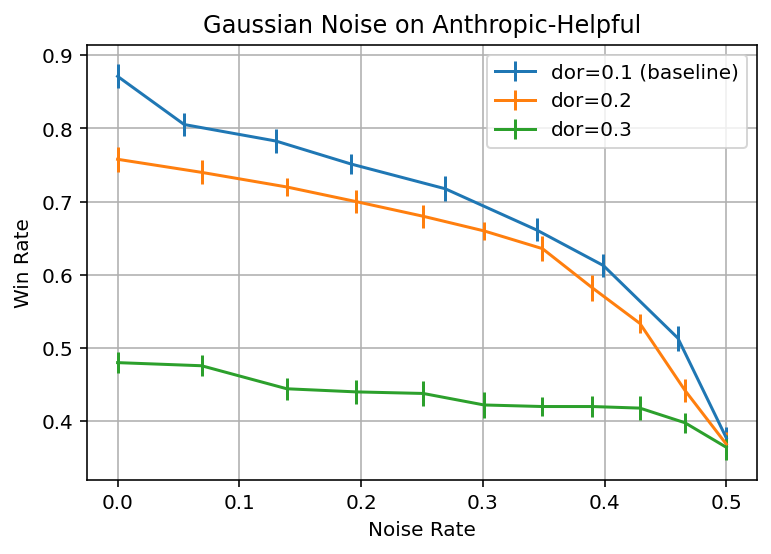}
\end{subfigure}
\caption{Alignment performance with different dropout rates (dor).}
\label{fig:results_do}
\end{figure}

Fig. \ref{fig:results_do} presents the alignment performance 
with different dropout rates.
We find that higher dropout rates significantly 
harm the alignment performance.
To summarize, our findings suggest that
high strength of regularization, in general,
fails to mitigate the negative impact of preference noise,
and in certain cases can even hurt the alignment performance.

\subsection{Data Filtering}
\label{subsec:data_filter}
Another approach to fight against noise is to
filter out noisy data in the 
Preference Pairs dataset (see Fig. \ref{fig:setup})
and only use the remaining data to perform alignment.
We use the popular 
\emph{confidence-based} data filtering method
\citep{cheng2008cleaning}.
For a prompt $x$ and two candidate responses $y_w, y_l$,
the \emph{confidence of $y_w$ preferred over $y_l$},
denoted $c(y_w > y_l | x)$, is a real value between 0 and 1. 
Ideally, the confidence function $c$ should be \emph{well-calibrated}
\citep{silva2023calibration},
i.e., $c(y_w > y_l | x)$ is identical to the true probability
of $y_w$ preferred over $y_l$.
In confidence-based data filtering, only pairs with the confidence level 
larger than a pre-defined threshold $t$ are used in training;
hence, if threshold $t=0.5$, no filtering is performed;
$t=1$ means all data will be removed.
We experiment with $t=0.5, 0.6, \cdots, 0.9$ to 
investigate the effect of different levels of data filtering.

Based on the user studies made by \citet{gao20irj}, 
we use the \citet{bradley1952rank} 
model to estimate the confidence value:
$c(y_w > y_l | x) = \sigma[r^*(y_w|x) - r^*(y_l|x)]$,
where $\sigma$ is the sigmoid function.
%
In practice, there are multiple 
methods for estimating the confidence values,
e.g., \emph{conformal predictors} 
\citep{shafer2008tutorial,einbinder2022conformaldl},
\emph{ensemble} methods \citep{liang2022rewarduncertainty},
and \emph{bias estimation} methods \citep{chen2023biasvariance}.

Note that with higher filtering threshold $t$,
the \emph{quality} of the remaining data is improved
at the cost of \emph{quantity} loss.
We deliberately do not back-fill the filtered data,
because in practice it can be prohibitively expensive
to collect more preference pairs.
%
This setup also allows us to study the trade-off
between data quality and data quantity in preference-based
alignment.
Fig. \ref{fig:threshold_data_percent} in Appendix 
\ref{appendix} shows how the
size of the remaining data shrinks with higher 
confidence thresholds.
We find that in both datasets, the data size drops quite quickly 
with the growth of $t$ values: 
Almost 20\% data are filtered as $t$ increases by 0.1. 

\begin{figure}
\centering
\begin{subfigure}[b]{0.32\textwidth}
 \centering
 \includegraphics[width=\textwidth]{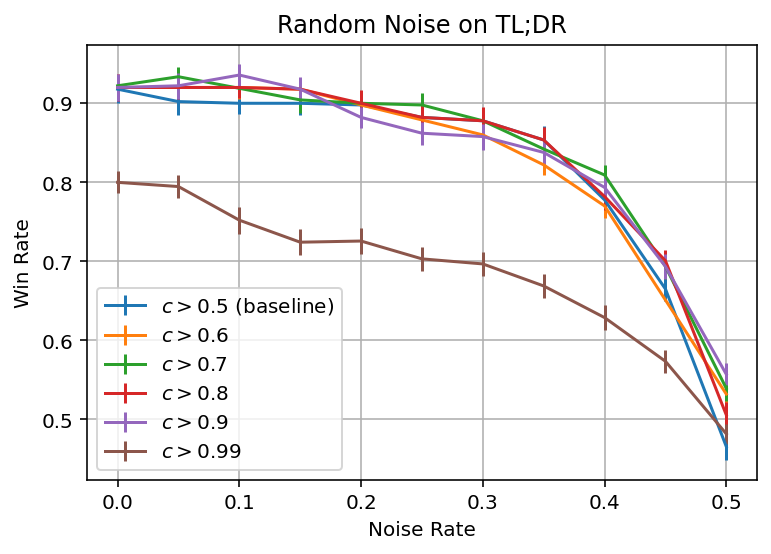}
\end{subfigure}
\begin{subfigure}[b]{0.32\textwidth}
 \centering
 \includegraphics[width=\textwidth]{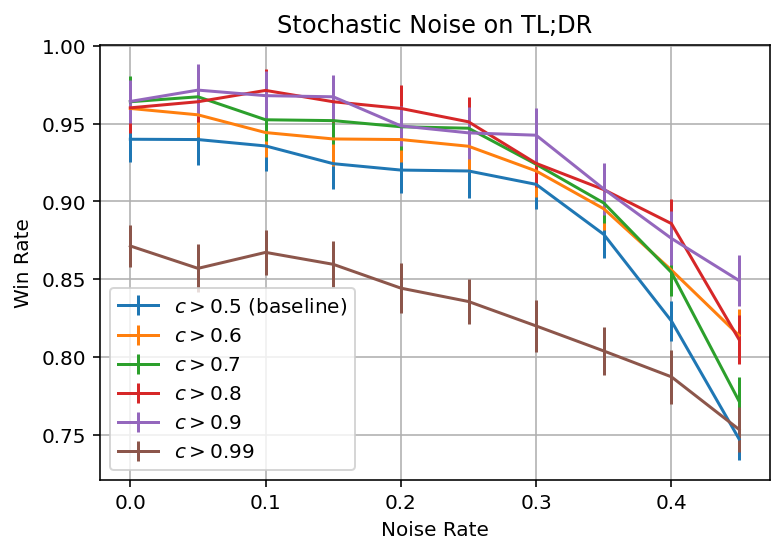}
\end{subfigure}
\begin{subfigure}[b]{0.32\textwidth}
 \centering
 \includegraphics[width=\textwidth]{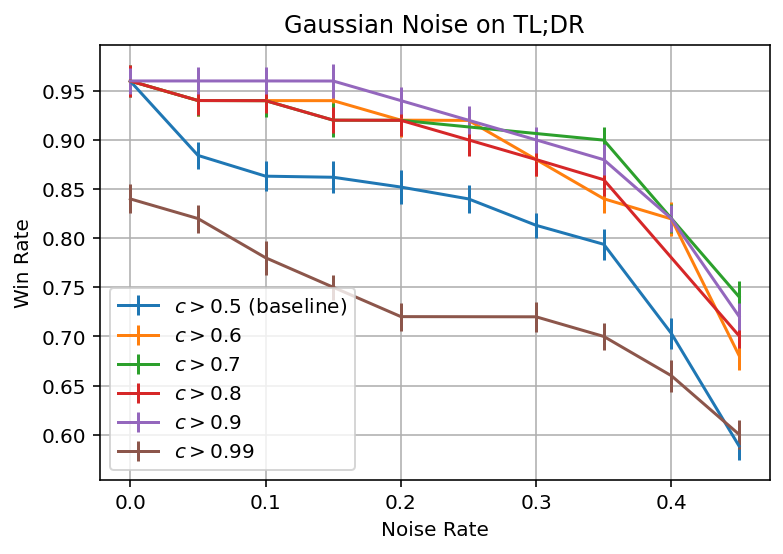}
\end{subfigure}
\begin{subfigure}[b]{0.32\textwidth}
 \centering
 \includegraphics[width=\textwidth]{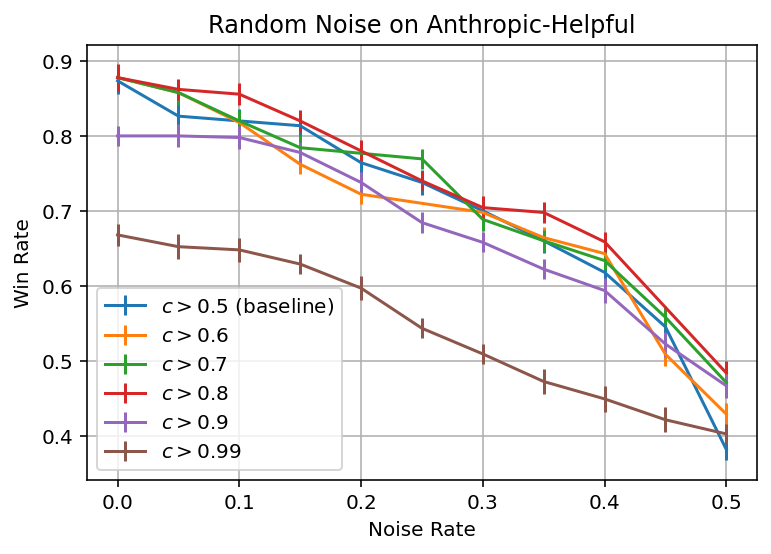}
\end{subfigure}
\begin{subfigure}[b]{0.32\textwidth}
 \centering
 \includegraphics[width=\textwidth]{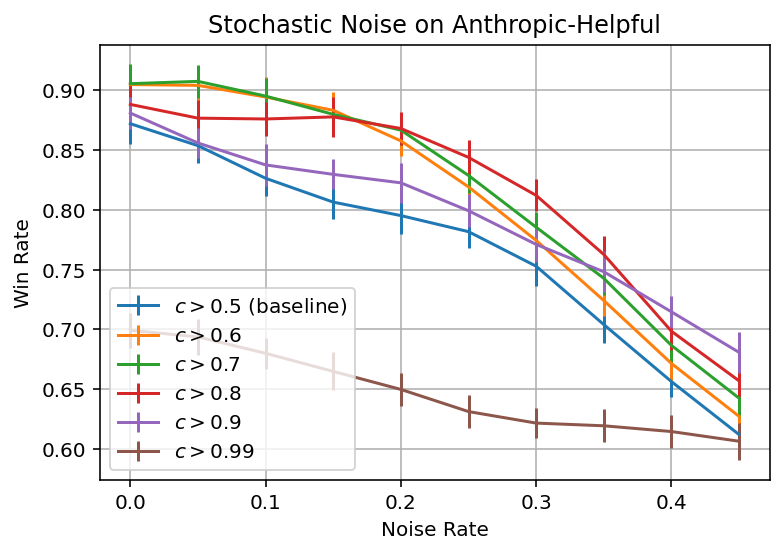}
\end{subfigure}
\begin{subfigure}[b]{0.32\textwidth}
 \centering
 \includegraphics[width=\textwidth]{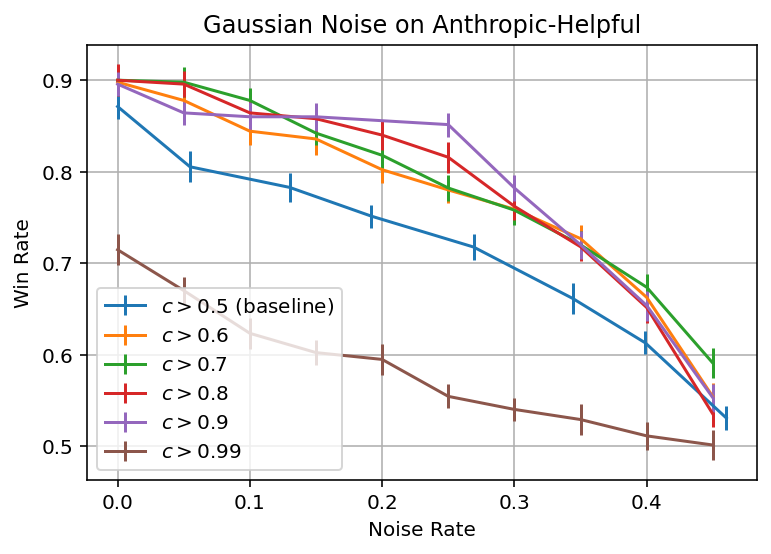}
\end{subfigure}
\caption{Alignment performance with confidence-based data filtering. 
}
\label{fig:data_filter_results}
\end{figure}

Fig. \ref{fig:data_filter_results} presents the alignment performance 
with different strengths of data filtering. 
We make the following observations.
\begin{itemize}
\item
\textbf{Data filtering does not help to fight against Random Noise.}
This is because Random Noise Oracle flips pairs
\emph{completely at random} \citep{frenay14noise}, 
i.e., the flipping
chance of each pair is uniformly at random, 
not affected by any other factors 
(e.g., the prompt $x$ or the responses $y_w, y_l$).
As a result, our confidence-based filtering cannot
reduce the number of the noisy pairs
in the filtered data, and hence fails to improve 
the performance.
\item 
\textbf{Data filtering is effective to 
mitigate the harm from Stochastic and Gaussian Noise.}
We find that with certain threshold (e.g., at $0.8$),
data filtering shows consistent and significant 
improvement across
all noise rates, noise types, and tasks. 
Considering that over 50\% preference pairs
have been removed with data filtering at
confidence threshold $0.8$, this result suggests
that data quality has a significant impact on the 
alignment performance.
\item
\textbf{Over-aggressive data filtering hurts the
performance.}
With very high confidence thresholds (e.g., $0.99$),
the alignment performance is compromised across
all noise rates, noise types, and tasks,
due to the severe loss of data size.
%
Hence, it is important
to properly trade off between data quality
and data quantity, in order to yield the optimal 
alignment performance.
\end{itemize}

%

To better understand how data filtering improves 
the data quality, we investigate the 
noise rate before and after data filtering, with
different noise types.
Fig. \ref{fig:filtered-noise-rate-appendix} in Appendix
\ref{appendix} shows the 
noise rates in data filtered with different confidence thresholds. 
We find that when the noise is from
Random Noise Oracle, the noise rate stays the same with
all data filtering thresholds; this explains why data
filtering does not help improve its performance.
When the noise is from Stochastic or Gaussian Oracles, 
data filtering can effectively reduce noise rate,
explaining the performance boost observed
in Fig. \ref{fig:data_filter_results}.


\section{Limitations \& Future Work}
\label{section:limitations}

\paragraph{Generalizability.}
We apply a popular alignment algorithm (DPO) to
T5-based language models in our experiments. 
We believe our observations presented in \S\ref{section:result_vanilla}
and \S\ref{section:result_mitigation} can be generalized to 
other alignment methods (e.g., PPO, SLiC and IPO,
because they essentially optimize the same objective function;
see \S\ref{section:related_works}) and other
Transformer \citep{Vaswani+2017} based language models, 
but this is yet to be empirically verified. 
Considering the large number of possible
combinations between alignment techniques and
language models, a systematic study is beyond the scope
of this paper, and we leave it for future work.

\paragraph{Mixture of Noise.}
We consider three strategies to add noise to the preferences,
but we only allow one strategy to be used at one time 
(see \S\ref{section:noisy_prefs}).
In practice, there may exist other types of noise, and
different types of noise can co-exist in preferences.
Our framework allows for creating new noise types by 
mixing primitive noise types,  but it remains
unclear whether the current observations can be applied
to the new noise types or not.
We hope our work can facilitate and encourage more work
on this direction.

\section{Conclusion}
\label{section:conclusion}
Pairwise preferences are widely used
for aligning Generative Language Models (GLMs)
with human values, but it remains unclear how
the \emph{noise in preferences} affect the
alignment performance, and how to mitigate
their negative impact.
To study these problems, we propose a framework
in which the types and rates of noise
can be simulated and controlled,
and we
%
perform systematic experiments
on two generation tasks (summary and dialogue generation)
with this framework.
Our findings suggest that
alignment performance can be highly sensitive
to the increase of noise rates, and 
appropriate data filtering is the most effective method to
mitigate the negative impact of noisy preferences.
Our work builds the first quantitative relation
between noise rates and alignment performance across
different noise types. 
We hope our work can help the community better understand
and mitigate the impact of preference noise
in GLM alignment.

\bibliography{colm2024_conference}

\begin{thebibliography}{44}
\providecommand{\natexlab}[1]{#1}
\providecommand{\url}[1]{\texttt{#1}}
\expandafter\ifx\csname urlstyle\endcsname\relax
  \providecommand{\doi}[1]{doi: #1}\else
  \providecommand{\doi}{doi: \begingroup \urlstyle{rm}\Url}\fi

\bibitem[Achiam et~al.(2023)Achiam, Adler, Agarwal, Ahmad, Akkaya, Aleman,
  Almeida, Altenschmidt, Altman, Anadkat, et~al.]{achiam2023gpt4}
Josh Achiam, Steven Adler, Sandhini Agarwal, Lama Ahmad, Ilge Akkaya,
  Florencia~Leoni Aleman, Diogo Almeida, Janko Altenschmidt, Sam Altman,
  Shyamal Anadkat, et~al.
\newblock Gpt-4 technical report.
\newblock \emph{arXiv preprint arXiv:2303.08774}, 2023.

\bibitem[Azar et~al.(2023)Azar, Rowland, Piot, Guo, Calandriello, Valko, and
  Munos]{azar2023ipo}
Mohammad~Gheshlaghi Azar, Mark Rowland, Bilal Piot, Daniel Guo, Daniele
  Calandriello, Michal Valko, and R{\'e}mi Munos.
\newblock A general theoretical paradigm to understand learning from human
  preferences.
\newblock \emph{arXiv preprint arXiv:2310.12036}, 2023.

\bibitem[Bai et~al.(2022{\natexlab{a}})Bai, Jones, Ndousse, Askell, Chen,
  DasSarma, Drain, Fort, Ganguli, Henighan, et~al.]{bai2022anthropichh}
Yuntao Bai, Andy Jones, Kamal Ndousse, Amanda Askell, Anna Chen, Nova DasSarma,
  Dawn Drain, Stanislav Fort, Deep Ganguli, Tom Henighan, et~al.
\newblock Training a helpful and harmless assistant with reinforcement learning
  from human feedback.
\newblock \emph{arXiv preprint arXiv:2204.05862}, 2022{\natexlab{a}}.

\bibitem[Bai et~al.(2022{\natexlab{b}})Bai, Kadavath, Kundu, Askell, Kernion,
  Jones, Chen, Goldie, Mirhoseini, McKinnon, et~al.]{bai2022constitutional}
Yuntao Bai, Saurav Kadavath, Sandipan Kundu, Amanda Askell, Jackson Kernion,
  Andy Jones, Anna Chen, Anna Goldie, Azalia Mirhoseini, Cameron McKinnon,
  et~al.
\newblock Constitutional ai: Harmlessness from ai feedback.
\newblock \emph{arXiv preprint arXiv:2212.08073}, 2022{\natexlab{b}}.

\bibitem[B{\"{o}}hm et~al.(2019)B{\"{o}}hm, Gao, Meyer, Shapira, Dagan, and
  Gurevych]{bohm18better}
Florian B{\"{o}}hm, Yang Gao, Christian~M. Meyer, Ori Shapira, Ido Dagan, and
  Iryna Gurevych.
\newblock Better rewards yield better summaries: Learning to summarise without
  references.
\newblock In Kentaro Inui, Jing Jiang, Vincent Ng, and Xiaojun Wan (eds.),
  \emph{Proceedings of the 2019 Conference on Empirical Methods in Natural
  Language Processing and the 9th International Joint Conference on Natural
  Language Processing, {EMNLP-IJCNLP} 2019, Hong Kong, China, November 3-7,
  2019}, pp.\  3108--3118. Association for Computational Linguistics, 2019.
\newblock \doi{10.18653/V1/D19-1307}.
\newblock URL \url{https://doi.org/10.18653/v1/D19-1307}.

\bibitem[Bradley \& Terry(1952)Bradley and Terry]{bradley1952rank}
Ralph~Allan Bradley and Milton~E Terry.
\newblock Rank analysis of incomplete block designs: I. the method of paired
  comparisons.
\newblock \emph{Biometrika}, 39\penalty0 (3/4):\penalty0 324--345, 1952.

\bibitem[Chen et~al.(2023)Chen, Lukasik, Jitkrittum, You, and
  Kumar]{chen2023biasvariance}
Lin Chen, Michal Lukasik, Wittawat Jitkrittum, Chong You, and Sanjiv Kumar.
\newblock It's an alignment, not a trade-off: Revisiting bias and variance in
  deep models.
\newblock \emph{arXiv preprint arXiv:2310.09250}, 2023.

\bibitem[Chen et~al.(2024)Chen, Deng, Yuan, Ji, and Gu]{chen2024selfplay}
Zixiang Chen, Yihe Deng, Huizhuo Yuan, Kaixuan Ji, and Quanquan Gu.
\newblock Self-play fine-tuning converts weak language models to strong
  language models.
\newblock \emph{arXiv preprint arXiv:2401.01335}, 2024.

\bibitem[Cheng et~al.(2008)Cheng, Chen, and Xie]{cheng2008cleaning}
Reynold Cheng, Jinchuan Chen, and Xike Xie.
\newblock Cleaning uncertain data with quality guarantees.
\newblock \emph{Proceedings of the VLDB Endowment}, 1\penalty0 (1):\penalty0
  722--735, 2008.

\bibitem[Cui et~al.(2023)Cui, Yuan, Ding, Yao, Zhu, Ni, Xie, Liu, and
  Sun]{cui2023ultrafeedback}
Ganqu Cui, Lifan Yuan, Ning Ding, Guanming Yao, Wei Zhu, Yuan Ni, Guotong Xie,
  Zhiyuan Liu, and Maosong Sun.
\newblock Ultrafeedback: Boosting language models with high-quality feedback.
\newblock \emph{arXiv preprint arXiv:2310.01377}, 2023.

\bibitem[Einbinder et~al.(2022)Einbinder, Romano, Sesia, and
  Zhou]{einbinder2022conformaldl}
Bat-Sheva Einbinder, Yaniv Romano, Matteo Sesia, and Yanfei Zhou.
\newblock Training uncertainty-aware classifiers with conformalized deep
  learning.
\newblock \emph{Advances in Neural Information Processing Systems},
  35:\penalty0 22380--22395, 2022.

\bibitem[Fr{\'{e}}nay \& Verleysen(2014)Fr{\'{e}}nay and
  Verleysen]{frenay14noise}
Beno{\^{\i}}t Fr{\'{e}}nay and Michel Verleysen.
\newblock Classification in the presence of label noise: {A} survey.
\newblock \emph{{IEEE} Trans. Neural Networks Learn. Syst.}, 25\penalty0
  (5):\penalty0 845--869, 2014.
\newblock \doi{10.1109/TNNLS.2013.2292894}.
\newblock URL \url{https://doi.org/10.1109/TNNLS.2013.2292894}.

\bibitem[Gao et~al.(2020)Gao, Meyer, and Gurevych]{gao20irj}
Yang Gao, Christian~M. Meyer, and Iryna Gurevych.
\newblock Preference-based interactive multi-document summarisation.
\newblock \emph{Inf. Retr. J.}, 23\penalty0 (6):\penalty0 555--585, 2020.
\newblock \doi{10.1007/S10791-019-09367-8}.
\newblock URL \url{https://doi.org/10.1007/s10791-019-09367-8}.

\bibitem[Gemma(2024)]{gemma24}
Gemma.
\newblock Gemma: Open models based on gemini research and technology.
\newblock Technical report, Google DeepMind, 2024.

\bibitem[Google(2023)]{team2023gemini}
Gemini~Team Google.
\newblock Gemini: A family of highly capable multimodal models.
\newblock \emph{arXiv preprint arXiv:2312.11805}, 2023.

\bibitem[Han et~al.(2020)Han, Yao, Liu, Niu, Tsang, Kwok, and
  Sugiyama]{DBLP:journals/corr/abs-2011-04406}
Bo~Han, Quanming Yao, Tongliang Liu, Gang Niu, Ivor~W. Tsang, James~T. Kwok,
  and Masashi Sugiyama.
\newblock A survey of label-noise representation learning: Past, present and
  future.
\newblock \emph{CoRR}, abs/2011.04406, 2020.
\newblock URL \url{https://arxiv.org/abs/2011.04406}.

\bibitem[H{\"u}llermeier \& Waegeman(2021)H{\"u}llermeier and
  Waegeman]{hullermeier2021aleatoric}
Eyke H{\"u}llermeier and Willem Waegeman.
\newblock Aleatoric and epistemic uncertainty in machine learning: An
  introduction to concepts and methods.
\newblock \emph{Machine Learning}, 110:\penalty0 457--506, 2021.

\bibitem[Jeon et~al.(2020)Jeon, Milli, and Dragan]{jeon2020reward}
Hong~Jun Jeon, Smitha Milli, and Anca Dragan.
\newblock Reward-rational (implicit) choice: A unifying formalism for reward
  learning.
\newblock \emph{Advances in Neural Information Processing Systems},
  33:\penalty0 4415--4426, 2020.

\bibitem[Ji et~al.(2023)Ji, Qiu, Chen, Zhang, Lou, Wang, Duan, He, Zhou, Zhang,
  et~al.]{ji2023aialignment}
Jiaming Ji, Tianyi Qiu, Boyuan Chen, Borong Zhang, Hantao Lou, Kaile Wang,
  Yawen Duan, Zhonghao He, Jiayi Zhou, Zhaowei Zhang, et~al.
\newblock Ai alignment: A comprehensive survey.
\newblock \emph{arXiv preprint arXiv:2310.19852}, 2023.

\bibitem[Kendall(1948)]{kendall1948rank}
Maurice~George Kendall.
\newblock Rank correlation methods.
\newblock 1948.

\bibitem[Kingsley \& Brown(2010)Kingsley and Brown]{kingsley2010preference}
David~C Kingsley and Thomas~C Brown.
\newblock Preference uncertainty, preference learning, and paired comparison
  experiments.
\newblock \emph{Land Economics}, 86\penalty0 (3):\penalty0 530--544, 2010.

\bibitem[Lee et~al.(2024)Lee, Phatale, Mansoor, Lu, Mesnard, Bishop, Carbune,
  and Rastogi]{lee2023rlaif}
Harrison Lee, Samrat Phatale, Hassan Mansoor, Kellie Lu, Thomas Mesnard, Colton
  Bishop, Victor Carbune, and Abhinav Rastogi.
\newblock Rlaif: Scaling reinforcement learning from human feedback with ai
  feedback.
\newblock In \emph{International Conference on Learning Representations}, 2024.

\bibitem[Lee et~al.(2021{\natexlab{a}})Lee, Smith, Dragan, and
  Abbeel]{lee2021bpref}
K~Lee, L~Smith, A~Dragan, and P~Abbeel.
\newblock B-pref: Benchmarking preference-based reinforcement learning.
\newblock \emph{Neural Information Processing Systems}, 2021{\natexlab{a}}.

\bibitem[Lee et~al.(2021{\natexlab{b}})Lee, Smith, and Abbeel]{lee2021pebble}
Kimin Lee, Laura Smith, and Pieter Abbeel.
\newblock Pebble: Feedback-efficient interactive reinforcement learning via
  relabeling experience and unsupervised pre-training.
\newblock \emph{arXiv preprint arXiv:2106.05091}, 2021{\natexlab{b}}.

\bibitem[Liang et~al.(2022)Liang, Shu, Lee, and
  Abbeel]{liang2022rewarduncertainty}
Xinran Liang, Katherine Shu, Kimin Lee, and Pieter Abbeel.
\newblock Reward uncertainty for exploration in preference-based reinforcement
  learning.
\newblock \emph{arXiv preprint arXiv:2205.12401}, 2022.

\bibitem[Lindner \& El-Assady(2022)Lindner and El-Assady]{lindner2022humans}
David Lindner and Mennatallah El-Assady.
\newblock Humans are not boltzmann distributions: Challenges and opportunities
  for modelling human feedback and interaction in reinforcement learning.
\newblock \emph{arXiv preprint arXiv:2206.13316}, 2022.

\bibitem[Liu et~al.(2024)Liu, Zhao, Joshi, Khalman, Saleh, Liu, and
  Liu]{liu2023statistical}
Tianqi Liu, Yao Zhao, Rishabh Joshi, Misha Khalman, Mohammad Saleh, Peter~J
  Liu, and Jialu Liu.
\newblock Statistical rejection sampling improves preference optimization.
\newblock In \emph{International Conference on Learning Representations}, 2024.

\bibitem[Munos et~al.(2023)Munos, Valko, Calandriello, Azar, Rowland, Guo,
  Tang, Geist, Mesnard, Michi, et~al.]{munos2023nash}
R{\'e}mi Munos, Michal Valko, Daniele Calandriello, Mohammad~Gheshlaghi Azar,
  Mark Rowland, Zhaohan~Daniel Guo, Yunhao Tang, Matthieu Geist, Thomas
  Mesnard, Andrea Michi, et~al.
\newblock Nash learning from human feedback.
\newblock \emph{arXiv preprint arXiv:2312.00886}, 2023.

\bibitem[Ouyang et~al.(2022)Ouyang, Wu, Jiang, Almeida, Wainwright, Mishkin,
  Zhang, Agarwal, Slama, Ray, et~al.]{ouyang2022}
Long Ouyang, Jeffrey Wu, Xu~Jiang, Diogo Almeida, Carroll Wainwright, Pamela
  Mishkin, Chong Zhang, Sandhini Agarwal, Katarina Slama, Alex Ray, et~al.
\newblock Training language models to follow instructions with human feedback.
\newblock \emph{Advances in Neural Information Processing Systems},
  35:\penalty0 27730--27744, 2022.

\bibitem[Rafailov et~al.(2023)Rafailov, Sharma, Mitchell, Ermon, Manning, and
  Finn]{rafailov2023dpo}
Rafael Rafailov, Archit Sharma, Eric Mitchell, Stefano Ermon, Christopher~D
  Manning, and Chelsea Finn.
\newblock Direct preference optimization: Your language model is secretly a
  reward model.
\newblock In \emph{Thirty-seventh Annual Conference on Neural Information
  Processing Systems}, 2023.

\bibitem[Raffel et~al.(2020)Raffel, Shazeer, Roberts, Lee, Narang, Matena,
  Zhou, Li, and Liu]{raffel2020t5x}
Colin Raffel, Noam Shazeer, Adam Roberts, Katherine Lee, Sharan Narang, Michael
  Matena, Yanqi Zhou, Wei Li, and Peter~J Liu.
\newblock Exploring the limits of transfer learning with a unified text-to-text
  transformer.
\newblock \emph{The Journal of Machine Learning Research}, 21\penalty0
  (1):\penalty0 5485--5551, 2020.

\bibitem[Schulman et~al.(2017)Schulman, Wolski, Dhariwal, Radford, and
  Klimov]{schulman2017ppo}
John Schulman, Filip Wolski, Prafulla Dhariwal, Alec Radford, and Oleg Klimov.
\newblock Proximal policy optimization algorithms.
\newblock \emph{arXiv preprint arXiv:1707.06347}, 2017.

\bibitem[Shafer \& Vovk(2008)Shafer and Vovk]{shafer2008tutorial}
Glenn Shafer and Vladimir Vovk.
\newblock A tutorial on conformal prediction.
\newblock \emph{Journal of Machine Learning Research}, 9\penalty0 (3), 2008.

\bibitem[Shazeer \& Stern(2018)Shazeer and Stern]{shazeer2018adafactor}
Noam Shazeer and Mitchell Stern.
\newblock Adafactor: Adaptive learning rates with sublinear memory cost.
\newblock In \emph{International Conference on Machine Learning}, pp.\
  4596--4604. PMLR, 2018.

\bibitem[Silva~Filho et~al.(2023)Silva~Filho, Song, Perello-Nieto,
  Santos-Rodriguez, Kull, and Flach]{silva2023calibration}
Telmo Silva~Filho, Hao Song, Miquel Perello-Nieto, Raul Santos-Rodriguez,
  Meelis Kull, and Peter Flach.
\newblock Classifier calibration: a survey on how to assess and improve
  predicted class probabilities.
\newblock \emph{Machine Learning}, 112\penalty0 (9):\penalty0 3211--3260, 2023.

\bibitem[Song et~al.(2022)Song, Kim, Park, Shin, and Lee]{song2022noisy}
Hwanjun Song, Minseok Kim, Dongmin Park, Yooju Shin, and Jae-Gil Lee.
\newblock Learning from noisy labels with deep neural networks: A survey.
\newblock \emph{IEEE Transactions on Neural Networks and Learning Systems},
  2022.

\bibitem[Stiennon et~al.(2020)Stiennon, Ouyang, Wu, Ziegler, Lowe, Voss,
  Radford, Amodei, and Christiano]{stiennon2020tldr}
Nisan Stiennon, Long Ouyang, Jeffrey Wu, Daniel Ziegler, Ryan Lowe, Chelsea
  Voss, Alec Radford, Dario Amodei, and Paul~F Christiano.
\newblock Learning to summarize with human feedback.
\newblock \emph{Advances in Neural Information Processing Systems},
  33:\penalty0 3008--3021, 2020.

\bibitem[Thurstone(2017)]{thurstone2017law}
Louis~L Thurstone.
\newblock A law of comparative judgment.
\newblock In \emph{Scaling}, pp.\  81--92. Routledge, 2017.

\bibitem[Touvron et~al.(2023)Touvron, Martin, Stone, Albert, Almahairi, Babaei,
  Bashlykov, Batra, Bhargava, Bhosale, et~al.]{touvron2023llama2}
Hugo Touvron, Louis Martin, Kevin Stone, Peter Albert, Amjad Almahairi, Yasmine
  Babaei, Nikolay Bashlykov, Soumya Batra, Prajjwal Bhargava, Shruti Bhosale,
  et~al.
\newblock Llama 2: Open foundation and fine-tuned chat models.
\newblock \emph{arXiv preprint arXiv:2307.09288}, 2023.

\bibitem[Vaswani et~al.(2017)Vaswani, Shazeer, Parmar, Uszkoreit, Jones, Gomez,
  Kaiser, and Polosukhin]{Vaswani+2017}
Ashish Vaswani, Noam Shazeer, Niki Parmar, Jakob Uszkoreit, Llion Jones,
  Aidan~N Gomez, \L~ukasz Kaiser, and Illia Polosukhin.
\newblock Attention is all you need.
\newblock In \emph{Advances in Neural Information Processing Systems},
  volume~30. Curran Associates, Inc., 2017.
\newblock URL
  \url{https://proceedings.neurips.cc/paper_files/paper/2017/file/3f5ee243547dee91fbd053c1c4a845aa-Paper.pdf}.

\bibitem[Yuan et~al.(2024)Yuan, Pang, Cho, Sukhbaatar, Xu, and
  Weston]{yuan2024selfreward}
Weizhe Yuan, Richard~Yuanzhe Pang, Kyunghyun Cho, Sainbayar Sukhbaatar, Jing
  Xu, and Jason Weston.
\newblock Self-rewarding language models.
\newblock \emph{arXiv preprint arXiv:2401.10020}, 2024.

\bibitem[Zhao et~al.(2023)Zhao, Joshi, Liu, Khalman, Saleh, and
  Liu]{zhao2023slichf}
Yao Zhao, Rishabh Joshi, Tianqi Liu, Misha Khalman, Mohammad Saleh, and Peter~J
  Liu.
\newblock Slic-hf: Sequence likelihood calibration with human feedback.
\newblock \emph{arXiv preprint arXiv:2305.10425}, 2023.

\bibitem[Zheng et~al.(2023)Zheng, Chiang, Sheng, Zhuang, Wu, Zhuang, Lin, Li,
  Li, Xing, Zhang, Gonzalez, and Stoica]{zheng2023llmjudge}
Lianmin Zheng, Wei-Lin Chiang, Ying Sheng, Siyuan Zhuang, Zhanghao Wu, Yonghao
  Zhuang, Zi~Lin, Zhuohan Li, Dacheng Li, Eric~P. Xing, Haotong Zhang, Joseph
  Gonzalez, and Ion Stoica.
\newblock Judging llm-as-a-judge with mt-bench and chatbot arena.
\newblock In \emph{Neural Information Processing Systems, Datasets and
  Benchmarks Track}, 2023.

\bibitem[Ziebart et~al.(2008)Ziebart, Maas, Bagnell, Dey,
  et~al.]{ziebart2008maximum}
Brian~D Ziebart, Andrew~L Maas, J~Andrew Bagnell, Anind~K Dey, et~al.
\newblock Maximum entropy inverse reinforcement learning.
\newblock In \emph{Aaai}, volume~8, pp.\  1433--1438. Chicago, IL, USA, 2008.

\end{thebibliography}
\bibliographystyle{colm2024_conference}

\appendix
\section{Appendix}
\label{appendix}

\begin{figure}[h]
\centering
\begin{subfigure}[b]{0.48\textwidth}
 \centering
 \includegraphics[width=\textwidth]{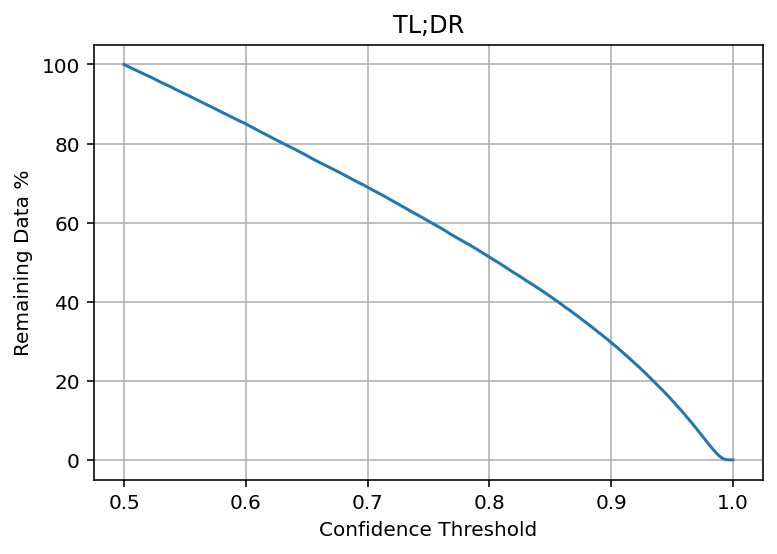}
\end{subfigure}
\begin{subfigure}[b]{0.48\textwidth}
 \centering
 \includegraphics[width=\textwidth]{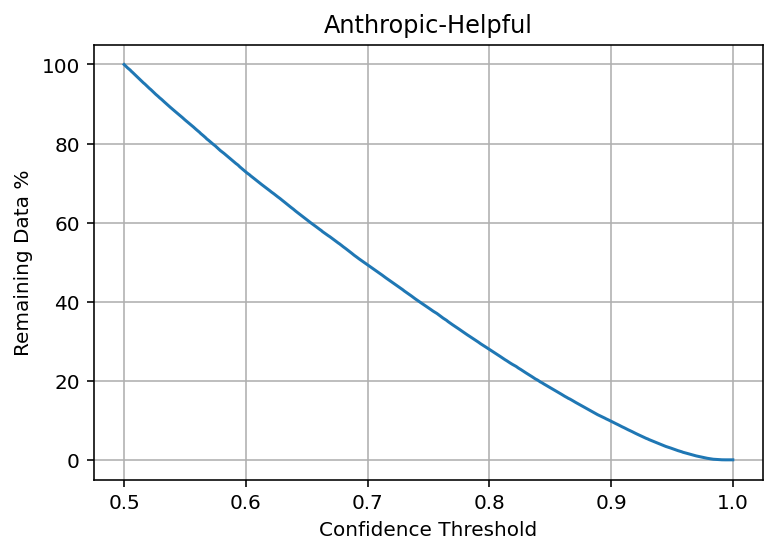}
\end{subfigure}
\caption{With higher data-selection threshold, fewer data will be 
remained for training.}
\label{fig:threshold_data_percent}
\end{figure}

\begin{figure}
\centering
\begin{subfigure}[b]{0.48\textwidth}
 \centering
 \includegraphics[width=\textwidth]{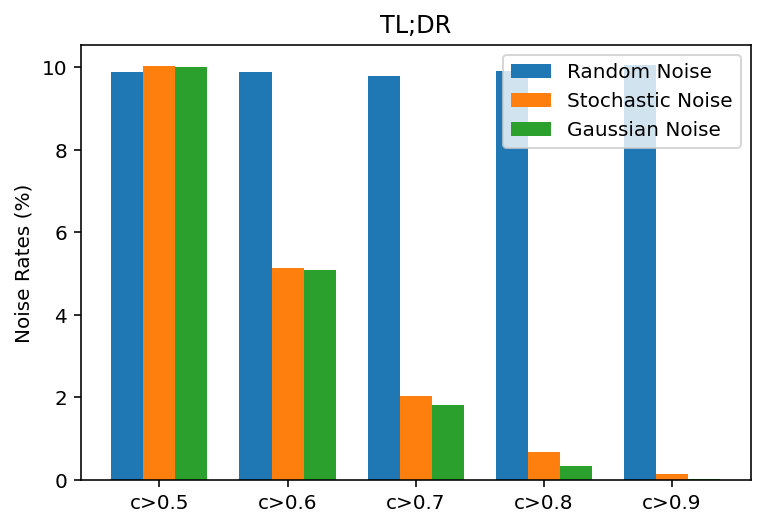}
\end{subfigure}
\begin{subfigure}[b]{0.48\textwidth}
 \centering
 \includegraphics[width=\textwidth]{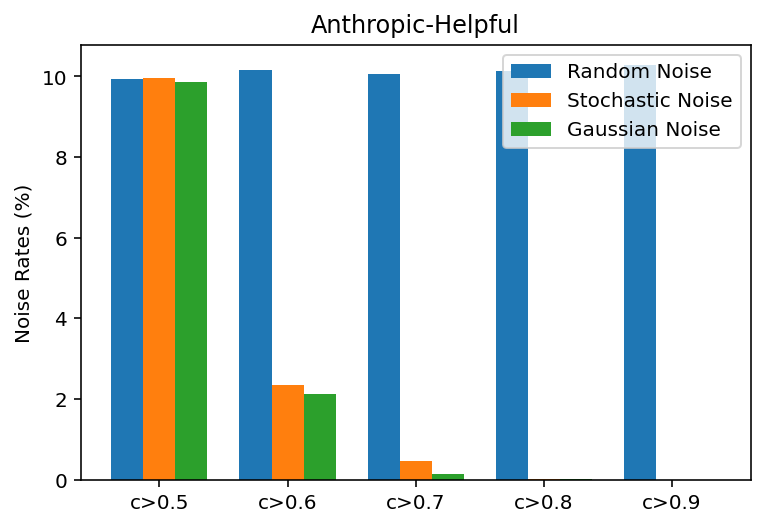}
\end{subfigure}
\begin{subfigure}[b]{0.48\textwidth}
 \centering
 \includegraphics[width=\textwidth]{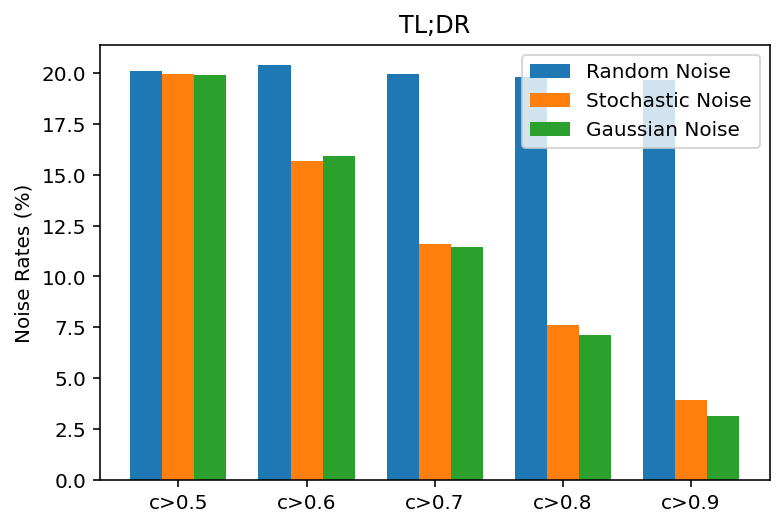}
\end{subfigure}
\begin{subfigure}[b]{0.48\textwidth}
 \centering
 \includegraphics[width=\textwidth]{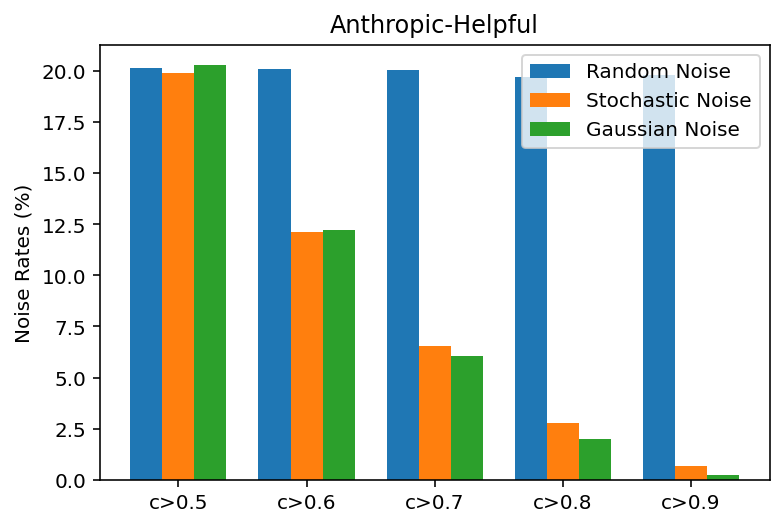}
%
\end{subfigure}
\begin{subfigure}[b]{0.48\textwidth}
 \centering
 \includegraphics[width=\textwidth]{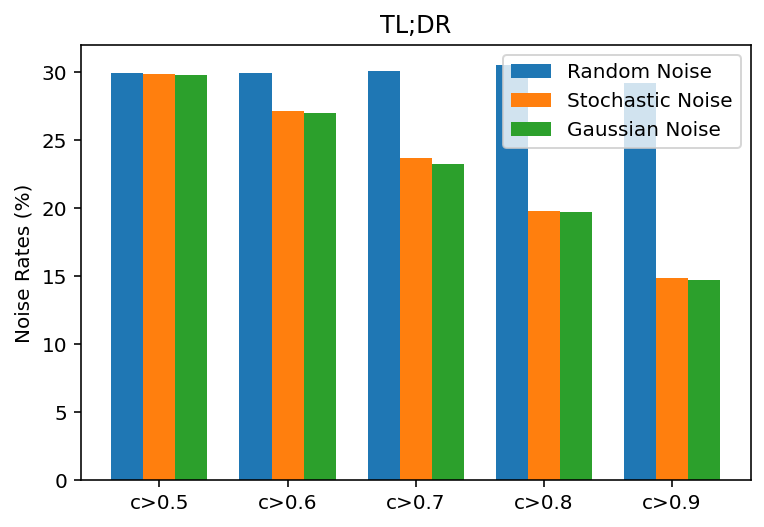}
\end{subfigure}
\begin{subfigure}[b]{0.48\textwidth}
 \centering
 \includegraphics[width=\textwidth]{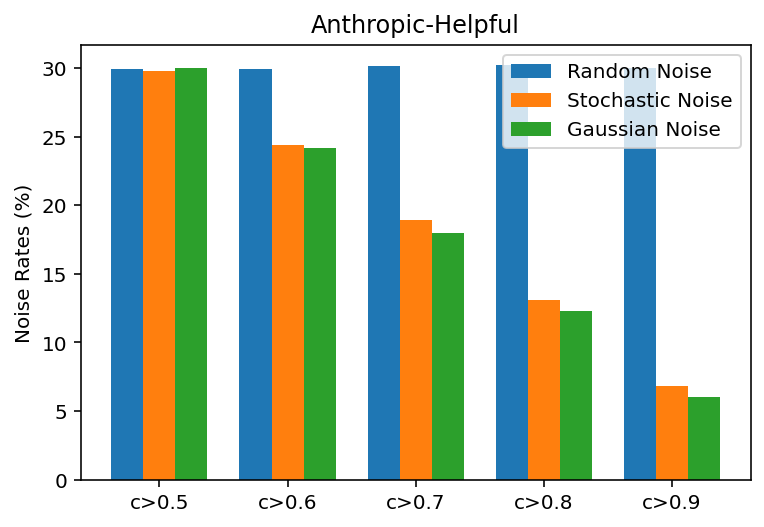}
\end{subfigure}
\begin{subfigure}[b]{0.48\textwidth}
 \centering
 \includegraphics[width=\textwidth]{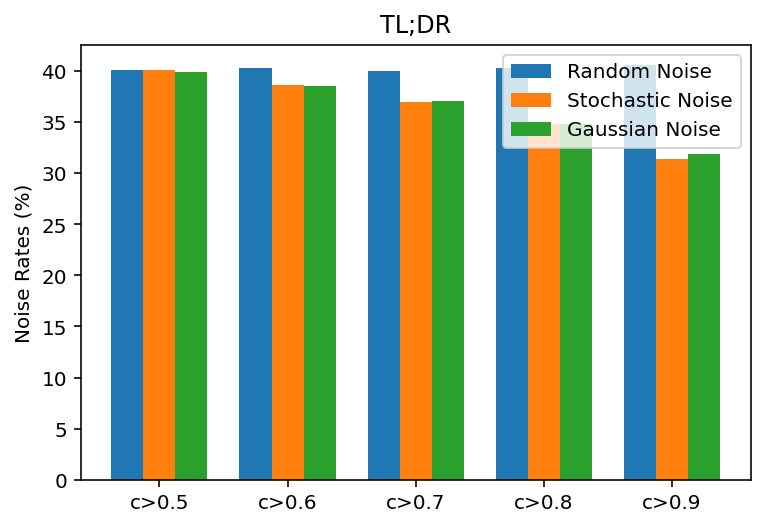}
\end{subfigure}
\begin{subfigure}[b]{0.48\textwidth}
 \centering
 \includegraphics[width=\textwidth]{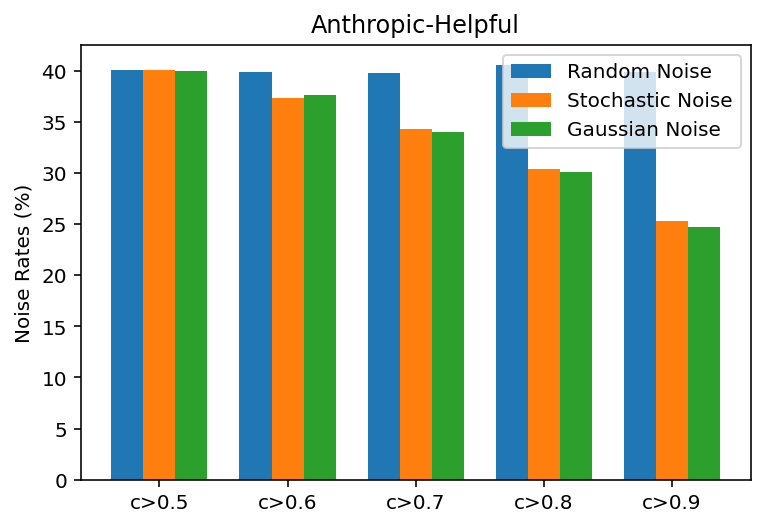}
\end{subfigure}
\caption{Noise rates of the filtered data, with different data filtering threshold. Note that when the threshold is $0.5$, no data is filtered. Here the original/unfiltered data has 10 - 40 \%
noise.
}
\label{fig:filtered-noise-rate-appendix}
\end{figure}

\end{document}